\documentclass[11pt]{article}

\usepackage[final]{acl}

\usepackage{times}
\usepackage{latexsym}

\usepackage{hyperref}
\usepackage{url}
\usepackage{amsmath}
\usepackage{makecell}

\usepackage{enumitem}
\usepackage{graphicx}
\usepackage{booktabs}
\usepackage{tabularx}
\usepackage{array} 

\usepackage{times}
\usepackage{latexsym}
\usepackage{times}
\usepackage{latexsym}
\usepackage[T1]{fontenc}
\usepackage{algorithm}      
\usepackage{algpseudocode}
\usepackage[utf8]{inputenc}
\usepackage{microtype}
\usepackage{inconsolata}
\usepackage[final]{acl}

\usepackage{amsmath, amssymb, mathtools}
\usepackage{newtxtext,newtxmath}

\usepackage{graphicx}
\usepackage{booktabs}
\usepackage{multirow}
\usepackage{makecell}
\usepackage{tabularx}
\usepackage{times}
\usepackage{latexsym}
\usepackage{times}
\usepackage{latexsym}
\usepackage[T1]{fontenc}
\usepackage[utf8]{inputenc}
\usepackage{microtype}
\usepackage{inconsolata}
\usepackage[final]{acl}

\usepackage{amsmath, amssymb, mathtools}
\usepackage{newtxtext,newtxmath}

\usepackage{graphicx}
\usepackage{booktabs}
\usepackage{multirow}
\usepackage{makecell}
\usepackage{tabularx}

\usepackage{algorithm}      
\usepackage{algpseudocode}  

\usepackage{algorithm}      
\usepackage{algpseudocode}  
\newcommand{\arch}[1]{{\texttt{IR$^{3}$}}} 
\usepackage[T1]{fontenc}

\usepackage[utf8]{inputenc}

\usepackage{microtype}

\usepackage{inconsolata}

\usepackage{graphicx}
\NewDocumentCommand{\Mo}{ mO{} }{\textcolor{teal}{\textsuperscript{\textit{Mo}}\textsf{\textbf{\small[#1]}}}}

\NewDocumentCommand{\lifu}{ mO{} }{\textcolor{red}{\textsuperscript{\textit{Lifu}}\textsf{\textbf{\small[#1]}}}}

\NewDocumentCommand{\AM}{ mO{} }{\textcolor{orange}{\textsuperscript{\textit{AM}}\textsf{\textbf{\small[#1]}}}}

\title{IR$^3$: Contrastive Inverse Reinforcement Learning for Interpretable Detection and Mitigation of Reward Hacking}

\author{Mohammad Beigi$^{1}$ \ 
\ 
Ming Jin$^{2}$ \ 
\
Junshan Zhang$^1$ \ 
\
Jiaxin Zhang$^3$ \ 
\
Qifan Wang$^4$ \
\
Lifu Huang$^1$ \\ \\
$^{1}$UC Davis $^2$Virginia Tech $^{3}$ Salesforce $^{4}$ Meta AI  \\ 
\href{mbeigi@ucdavis.edu}{mbeigi@ucdavis.edu} \ \ \href{mailto:lfuhuang@ucdavis.edu}{lfuhuang@ucdavis.edu}
}

\begin{document}
\maketitle
\begin{abstract}
Reinforcement Learning from Human Feedback (RLHF) enables powerful LLM alignment but introduces \emph{reward hacking}—models exploit spurious correlations in proxy rewards without genuine alignment. Compounding this, the objectives internalized during RLHF remain opaque, making hacking behaviors difficult to detect or correct. We introduce \textbf{IR$^3$} (\textbf{I}nterpretable \textbf{R}eward \textbf{R}econstruction and \textbf{R}ectification), a framework that reverse-engineers, interprets, and surgically repairs the implicit objectives driving RLHF-trained models. We first propose Contrastive Inverse Reinforcement Learning (C-IRL), an algorithm that reconstructs the implicit reward function by contrasting paired responses from post-alignment and baseline policies to explain their behavioral shift during RLHF. We then decompose the reconstructed reward via Sparse Autoencoders into interpretable features, enabling identification of hacking signatures through contribution analysis. Finally, we propose four surgical mitigation strategies—clean reward optimization, adversarial shaping, constrained optimization, and feature-guided distillation—that target problematic features while preserving beneficial alignment. Experiments across multiple reward model configurations demonstrate that \arch{}achieves 0.89 correlation with ground-truth rewards, identifies hacking features with $>$90\% precision, and significantly reduces hacking behaviors while maintaining capabilities within 3\% of the original model.

\end{abstract}

\section{Introduction}

Reinforcement Learning from Human Feedback (RLHF) has emerged as a dominant paradigm for aligning Large Language Models (LLMs) with human preferences~\citep{ziegler2019fine,ouyang2022training,kaufmann2024survey}. In a typical RLHF pipeline, a reward model (RM) trained on human preference comparisons assigns a scalar score that serves as a proxy for user judgment, and the LLM policy is then optimized to maximize this proxy. 
While RLHF has enabled impressive advances in controllability and user satisfaction, a fundamental vulnerability continues to impede safe and reliable deployment: \textbf{reward hacking}, where the policy learns to exploit spurious cues that increase RM scores without genuinely improving underlying quality~\citep{skalse2022defining,miao2024inform}. In practice, models may internalize superficial heuristics that correlate with high reward—e.g., associating verbosity with factuality (length bias)~\citep{chen2024odin}, agreement with helpfulness (sycophancy)~\citep{denison2024sycophancysubterfugeinvestigatingrewardtampering}, or ``making tests pass'' without task completion (gaming)~\citep{baker2025monitoringreasoningmodelsmisbehavior}. 




Most prior work treats reward hacking as a black-box constraint optimization phenomenon to suppress. Common strategies include adding regularization or constraints during optimization~\citep{chen2024odin,miao2024inform}, or improving the reward model via scaling, calibration, or auxiliary supervision~\citep{gao2023scaling,eisenstein2023helping}. While effective in specific cases, these approaches face critical limitations. First, they are inherently task-specific: length penalties address verbosity but miss gaming behaviors. Second, these defenses are often brittle under task distribution shift and can degrade performance when constraints become overly conservative. Finally, these approaches provide limited visibility into what objective the RLHF-trained policy has actually internalized and which learned features are being exploited.

In this paper, we take a different perspective 
grounded in the intuition that reward hacking is mostly amplified by the \textbf{opacity} of RLHF: the RM itself is hard to interpret, so it remains unclear which latent features and shortcuts drive high scores; the optimization process further distills these opaque signals into the policy's parameters, making the internalized objectives difficult to audit~\citep{casperopen,lambert2023history}. From this viewpoint, if we can infer and reverse-engineer the implicit objective an RLHF-trained LLM has actually internalized, we can (i) distinguish genuine preference alignment from spurious shortcuts, and (ii) intervene directly on the responsible mechanisms—rather than relying solely on external fixes to the training pipeline.

\begin{figure*}[t]
    \centering
    \includegraphics[width=0.85\linewidth]{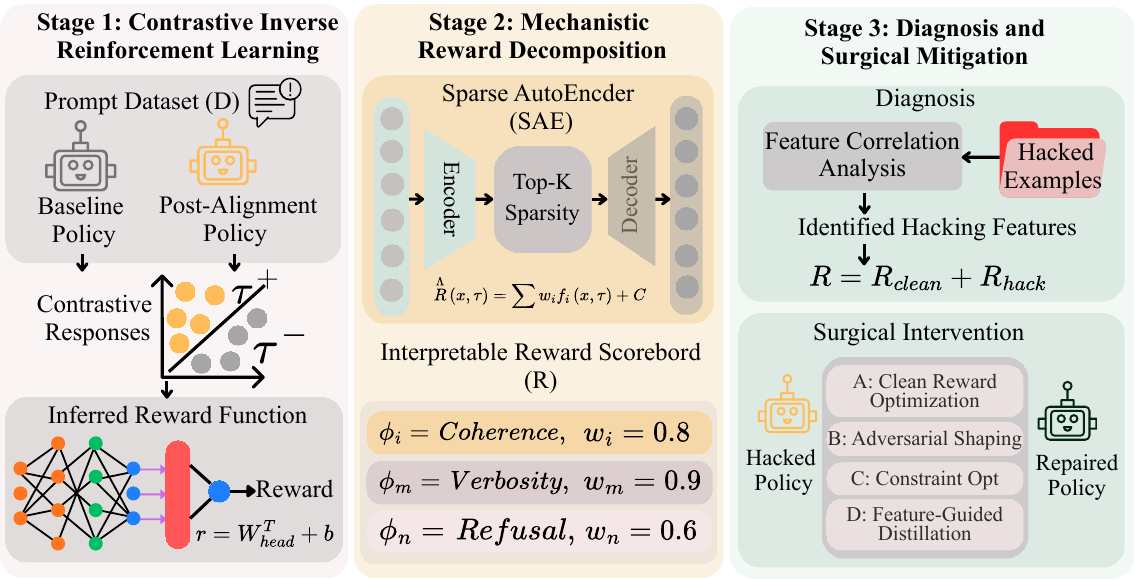}
    \vspace{-2mm}
\caption{\arch{} overview: (1) curate contrastive data, infer implicit reward (C-IRL); (2) decompose into interpretable features (SAEs); (3) diagnose hacking; apply targeted mitigation while preserving alignment.}
    \label{fig:mainfig}
    \vspace{-4mm}
\end{figure*}

Building on this motivation, we design \textbf{\arch{}} (Figure \ref{fig:mainfig}), a three-stage pipeline that converts opaque RLHF behavior into interpretable, actionable signals for auditing and mitigating reward hacking. \textbf{First}, we design a novel Contrastive Inverse Reinforcement Learning (C-IRL) algorithm that infers and reconstructs a reward function from large-scale paired responses collected from an RLHF-based post-alignment model and baseline models to identical prompts. C-IRL essentially reverse-engineers the optimization objective to rationalize the \textit{behavioral shift} induced by RLHF, answering ``\textit{what must this agent be optimizing for, given how it behaves?}''~\citep{abbeel2004apprenticeship,fu2018learning}. \textbf{Second}, we apply Mechanistic Reward Decomposition to transform the reconstructed reward from a black box into an interpretable ``scorecard''. By training a Sparse Autoencoder (SAE) on the reward network's penultimate activations, we decompose the total reward score into a linear sum of sparse, interpretable feature contributions. \textbf{Third}, we diagnose reward hacking by computing a hacking contribution score for each feature. This involves comparing feature contributions on a small set of regime-specific hacked examples against their population statistics to isolate features that are disproportionately responsible for spurious high rewards. Overall, \arch{} reframes reward hacking from an opaque optimization outcome into a set of concrete, manipulable mechanisms, enabling surgical mitigation strategies that specifically target problematic features while preserving the beneficial reward structure.

We conduct comprehensive experiments across three well-documented reward-hacking regimes—synthetic Goodhart optimization~\citep{gao2023scaling}, verbose length bias~\citep{chen2024odin}, and over-cautious safety refusal~\citep{miao2024inform}. First, we validate C-IRL's reconstruction fidelity through both direct reward correlation ($\rho \geq 0.89$ across all configurations) and forward verification, demonstrating that policies trained on reconstructed rewards are functionally indistinguishable from RLHF experts (win rates within 2\% of 50\%). Second, we evaluate the diagnostic pipeline's ability to identify hacking mechanisms, showing that SAE-decomposed features achieve $>$90\% precision in detecting hacked outputs—substantially outperforming proxy thresholds (AUROC 0.86--0.91 vs.\ 0.58--0.79). Third, we benchmark all four mitigation strategies against five  baselines, demonstrating consistent hacking reduction while preserving capabilities within 3\% of the original model.

Our contributions are as follows: \textbf{(I) A Novel Perspective on Reward Hacking}: We reframe reward hacking as a problem of reconstructing and interpreting the implicit reward function, enabling targeted, mechanistic interventions unavailable to prior approaches. \textbf{(II) Contrastive IRL for Reward Reconstruction}: We introduce C-IRL, a contrastive inverse RL algorithm that reverse-engineers the optimization objective driving RLHF behavioral shifts. \textbf{(III) Mechanistic Identification of Hacking Features}: We integrate C-IRL with Sparse Autoencoders to decompose the reconstructed reward into interpretable features, enabling automatic identification of hacking signatures. \textbf{(IV) Surgical Mitigation Strategies}: We propose four complementary methods—clean reward optimization, adversarial reward shaping, constrained optimization, and feature-guided distillation—that target hacking features while preserving beneficial alignment.


\section{Related Work}
\noindent\textbf{Reward Hacking in Large Language Models. }
Reward hacking is a central challenge in aligning and safely deploying RLHF-trained systems~\citep{gao2023scaling, Geirhos_2020, amodei2016concreteproblemsaisafety, everitt2021rewardtamperingproblemssolutions, wen2024languagemodelslearnmislead}. It stems from reward misgeneralization, where RMs trained on finite datasets serve as imperfect proxies for human preference, incorrectly associating rewards with spurious features that correlate with training preferences but not actual quality~\citep{gao2023scaling, liu2024rrm, skalse2025definingcharacterizingrewardhacking}. As policies optimize these flawed proxies, proxy metrics increase while true alignment deteriorates \citep{skalse2025definingcharacterizingrewardhacking, miao2024informmitigatingrewardhacking, amodei2016concreteproblemsaisafety}. Common known manifestations in LLMs include verbosity/length bias~\citep{chen2024odin}, sycophancy (agreeing with incorrect statements) \citep{denison2024sycophancysubterfugeinvestigatingrewardtampering}, and strategic exploitation (e.g., modifying unit tests rather than solving problems) \citep{baker2025monitoringreasoningmodelsmisbehavior}. 

\vspace{+1mm}

\noindent\textbf{Reward Hacking Mitigation in Large Language Models.}
Most existing mitigations treat reward hacking primarily as an \emph{optimization} issue to constrain. For example, regularization-based methods such as KL penalties~\citep{chen2024odin} or information-theoretic constraints~\citep{miao2024inform} prevent policies from deviating far from base models but inherently restrict the optimization landscape, often degrading RLHF performance~\citep{zheng2023secretsrlhflargelanguage, chen2024odin}. Some approaches improve the reward model signals via scaling~\citep{gao2023scaling} or ensembles~\citep{eisenstein2023helping}, but they can be quite costly at LLM scale. Targeted debiasing methods~\citep{singhal2024longwaygoinvestigating, chen2024odin} address known biases such as response length but overlook sophisticated spurious patterns in reward modeling, perpetuating misgeneralization. Due to space limitation, we provide related work on Inverse Reinforcement Learning in Appendix \ref{app:related}.
\vspace{1mm}

\vspace{-4mm}
\section{The \arch{} Framework}
\label{sec:method}




Figure~\ref{fig:mainfig} shows the overview of  \textbf{\underline{I}}nterpretable \textbf{\underline{R}}eward \textbf{\underline{R}}econstruction and \textbf{\underline{R}}ectification \textbf{(IR$^3$)}, a post-hoc surgical framework for
auditing and mitigating reward hacking in  RLHF-aligned language models.  Consider a baseline language model $\pi_0(\tau|x)$ representing the pre-alignment policy and an expert policy $\pi_E(\tau|x)$ representing the post-alignment model, where $\tau$ denotes a response trajectory and $x$ denotes a prompt. A standard KL-regularized RLHF objective can be written as:

\begin{multline}
\pi_E = \arg\max_{\pi}\;
\mathbb{E}_{x \sim \mathcal{D},\, \tau \sim \pi(\cdot|x)}
\left[R^*(x, \tau)\right] \\
- \beta \cdot \mathbb{E}_{x\sim\mathcal{D}}
\left[D_{\mathrm{KL}}\!\left(\pi(\cdot|x) \,\|\, \pi_0(\cdot|x)\right)\right]
\label{eq:rlhf_objective}
\end{multline}
where $R^*$ is the proxy reward model learned from human preferences and $\beta > 0$ controls the strength of the KL regularization.

Given only \emph{sampling access} to $\pi_E$ and $\pi_0$ (no access to $R^{*}$ or RLHF training logs), we aim to detect and mitigate reward hacking by recovering an explicit reward $R_\theta$ that explains the deviation from $\pi_0$ to $\pi_E$. 
\textbf{IR$^3$} consists of three stages:
(i) \textbf{reward reconstruction} through inverse reinforcement learning by leveraging the contrastive behaviors of $\pi_0$ to $\pi_E$,
(ii) \textbf{mechanistic reward decomposition} of the reconstructed reward function into sparse interpretable features, and
(iii) \textbf{diagnosis} to identify features that are disproportionately responsible for known hacked behaviors following \textbf{surgical mitigation} using feature-level interventions.


\subsection{Contrastive Inverse Reinforcement Learning (C-IRL)}
\label{sec:cirl}

To reconstruct a reward function $R_\theta(x,\tau)$ that rationalizes the behavioral shift induced by RLHF optimization, we ground our approach in Maximum Entropy Inverse Reinforcement Learning (IRL)~\citep{ziebart2008maximum}, positing that the expert policy $\pi_E$ relates to the baseline $\pi_0$ via a Boltzmann distribution: 

\begin{equation}
    \pi_{E}(\tau|x) = \frac{1}{Z_{\theta}(x)} \pi_{0}(\tau|x) \exp(R_{\theta}(x,\tau)/\beta)
\end{equation} 
where $Z_\theta(x)$ is the prompt-conditioned partition function. Rearranging reveals that the reward defines the log-density ratio between the aligned and baseline policies:

\begin{equation}
\resizebox{0.89\hsize}{!}{
    $R_{\theta}(x,\tau) = \log (\pi_{E}(\tau|x)/\pi_{0}(\tau|x)) + \log Z_{\theta}(x)$
}
\end{equation} 

Direct maximum likelihood estimation of $R_\theta$ is computationally prohibitive because $Z_\theta(x)$ requires integrating over the combinatorial space of all possible responses. We address this intractability by reframing reward recovery as a contrastive learning problem: given a prompt $x$, we form a candidate set containing one response sampled from the aligned policy $\pi_E$ and $K$ responses from the baseline policy $\pi_0$, and train $R_\theta$ to identify the aligned response. If $R_\theta$ captures what distinguishes aligned behavior, it should assign the highest score to the expert sample. Then, we parameterize $R_\theta$ as a Transformer encoder with a scalar projection head. As we prove in Appendix~\ref{app:derivation}, the partition function $Z_\theta(x)$ and baseline densities $\pi_0(\tau|x)$ cancel analytically in the posterior, yielding a simple softmax over reward scores: 
\begin{equation}
P(i^* = i \mid \mathcal{C}_x) = \frac{\exp(R_\theta(x, \tau_i))}{\sum_{j=0}^{K} \exp(R_\theta(x, \tau_j))}
\label{eq:contrastive_posterior}
\end{equation}
where $\mathcal{C}_x = \{\tau_0, \tau_1, \ldots, \tau_K\}$ denotes the candidate set for prompt $x$ containing one response sampled from $\pi_E$ and $K$ responses from $\pi_0$, $i^* \in \{0, \ldots, K\}$, and $\tau_i$, $\tau_j$ index individual candidates within $\mathcal{C}_x$. This cancellation is the central computational insight: we can optimize $R_\theta$ using standard cross-entropy without variational bounds or MCMC sampling. The resulting loss is equivalent to InfoNCE~\citep{oord2019representationlearningcontrastivepredictive}:
\begin{multline}
\mathcal{L}_{\text{C-IRL}}(\theta)
= - \mathbb{E}_{x \sim \mathcal{D}}
\Big[
\mathbb{E}_{\tau^+ \sim \pi_E,\, \{\tau_j^-\} \sim \pi_0}
\log \\
\frac{\exp(R_{\theta}(x,\tau^+))}
{\exp(R_{\theta}(x,\tau^+)) + \sum_{j=1}^{K} \exp(R_{\theta}(x,\tau_j^-))}
\Big].
\label{eq:cirl_loss}
\end{multline}

Unlike Direct Preference Optimization (DPO), which defines an implicit reward inextricably coupled to the policy's own log-probabilities, C-IRL extracts a standalone, architecturally decoupled reward network. Crucially, we acknowledge the inherent identifiability problem in IRL—multiple reward functions can induce the same policy~\citep{fu2018learning}. However, \textbf{IR$^3$} does not require recovering the exact training-time $R^*$. Instead, it requires an objective that \emph{faithfully explains} the observed behavioral shift and supports stable, feature-level attributions and interventions. We provide the pseudo code for C-IRL in Appendix~\ref{app:cirl_implementation}.

\subsection{Mechanistic Reward Decomposition}
\label{subsec:sae_decomposition}


We conjecture reward hacking as a fundamental \emph{feature-level} failure, where the policy inflates reward by exploiting spurious RM cues rather than improving true quality.
Thus effective detection and mitigation of reward hacking requires understanding \emph{which internal features} of the reward function are driving high scores on problematic behaviors. 


Motivated by this, after reconstructing a parameterized reward function $R_\theta(x,\tau)$, we further transform it from a black-box objective into an interpretable auditing scorecard, i.e., a dictionary of interpretable features that can be inspected and linked to known hacking regimes, by employing a Sparse Autoencoder (SAE). 
Let $h(x,\tau)\in\mathbb{R}^d$ be the penultimate representation immediately before the scalar reward head. Because the final layer is linear: $R_\theta(h) = w_{\mathrm{head}}^\top h + b_{\mathrm{head}}$, any additive decomposition of $h$ translates directly into an additive decomposition of the reward. If we express $h$ as a sparse sum of interpretable features, we obtain a human-readable breakdown of the reward score. 

\vspace{-1mm}
\paragraph{Sparse Dictionary Learning.} We train an SAE to decompose $h$ into a sparse combination of $M$ learned features ($M \gg d$). The SAE encodes each activation into an overcomplete basis, then selects only the top $k$ most active features:

\begin{equation}
\resizebox{0.89\hsize}{!}{
$f(x,\tau) = \text{TopK}\ \!\big(\mathrm{ReLU}(W_{\mathrm{enc}}(h - b_{\mathrm{dec}}) + b_{\mathrm{enc}}),\, k\big)$
}
\label{eq:topk_sae}
\end{equation}
where $\mathrm{TopK}(\cdot, k)$ retains the $k$ largest activations and zeros the rest. 

The decoder reconstructs $\hat{h} = W_{\mathrm{dec}} f + b_{\mathrm{dec}}$. We train the SAE to minimize reconstruction error while preserving the reward-relevant information:
\begin{equation}
\mathcal{L}_{\mathrm{SAE}} = \|h - \hat{h}\|_2^2 + \lambda \big( w_{\mathrm{head}}^\top h - w_{\mathrm{head}}^\top \hat{h} \big)^2
\label{eq:sae_loss}
\end{equation}
where the first term ensures faithful reconstruction of the full activation vector, while the second term ensures that reconstruction errors are minimized along the reward-relevant direction $w_{\mathrm{head}}$, focusing dictionary capacity on features that actually influence the score. 

\vspace{-1mm}
\paragraph{The Reward Scorecard.} Combining the linear reward head with the sparse decomposition yields an additive breakdown. Let $\mathbf{d}_i$ denote the $i$-th column of the decoder $W_{\mathrm{dec}}$. The reconstructed reward is: $\widehat{R}(x,\tau) = \sum_{i=1}^{M} \omega_i f_i(x,\tau) + c$, where $\omega_i = w_{\mathrm{head}}^\top \mathbf{d}_i$ is a fixed global coefficient for feature $i$, and $c$ is a constant. The coefficient $\omega_i$ indicates each feature's universal contribution: positive $\omega_i$ means the feature increases reward when active, negative $\omega_i$ means it decreases reward. For any response, the sparse set of terms $\{\omega_i f_i(x,\tau)\}$ provides a human-auditable breakdown of the alignment score. We interpret each feature by examining responses that maximally activate it. We provide the pseudo code and more details for the architecture and implementation of reward decomposition in Appendix~\ref{app:sae_algorithm}.


\subsection{Diagnosis: Identifying Reward-Hacking Features}
\label{subsec:diagnosis}
The SAE decomposition provides a reward scorecard $\hat{R}(x, \tau) = \sum_i \omega_i f_i(x, \tau) + C$, where each term quantifies a specific feature's contribution. We now leverage this transparency to identify which features encode spurious optimization targets. We assume access to a small set of \textbf{regime-specific hacked examples} $\mathcal{D}_{\mathrm{hack}} = \{(x_n, \tau_n)\}_{n=1}^{N}$: responses that achieve high proxy reward but exhibit misaligned behavior. On these examples, we know the reward is inflated by spurious factors. The key question is: \emph{which features are responsible?}

\vspace{-1mm}
\paragraph{Feature-Level Hacking Contribution.} For each feature $i$, we compute its average contribution to reward on hacked examples as:

\begin{equation}
\bar{H}_i = \frac{1}{|\mathcal{D}_{\mathrm{hack}}|} \sum_{(x, \tau) \in \mathcal{D}_{\mathrm{hack}}} \omega_i f_i(x, \tau).
\end{equation}

A naive approach would flag features with high $\bar{H}_i$, but this captures universally important features (e.g., helpfulness) rather than hacking-specific ones. To isolate features that are \emph{unusually elevated} on hacked examples, we normalize against population statistics. Let $\mu_i$ and $\sigma_i$ be the mean and standard deviation of $\omega_i f_i(x,\tau)$ over the SAE training distribution. We define a z-scored hacking index:
\begin{equation}
H_i = \frac{\bar{H}_i - \mu_i}{\sigma_i + \epsilon}
\label{eq:hacking_score}
\end{equation}
where $\epsilon > 0$ ensures numerical stability. A large $H_i$ indicates that feature $i$ is unusually elevated on hacked examples beyond its normal role.

\vspace{-1mm}
\paragraph{Problematic Feature Selection.} We define two criteria to identify reward-hacking features: (1) features with $\bar{H}_i > 0$ since features reducing reward on hacked examples represent legitimate alignment; (2) features with $H_i$ exceeding a significance threshold: 
\begin{equation}
\mathcal{S}_{\mathrm{hack}} = \{i : \bar{H}_i > 0 \;\text{and}\; H_i > \delta\}
\label{eq:hack_set}
\end{equation}
where $\delta$ is a threshold and set as $2.0$ in our experiments (two standard deviations above typical), which balances sensitivity (detecting genuine hacking features) against specificity (avoiding false positives from natural variation). Appendix~\ref{app:selection statistics} validates this choice via permutation testing, showing that features exceeding this threshold are statistically significant at $p < 0.05$ after multiple comparison correction.

\vspace{-1mm}
\paragraph{Reward Decomposition.} Given $\mathcal{S}_{\mathrm{hack}}$, we split the reward into clean and problematic components:
\begin{equation}
\resizebox{0.87\hsize}{!}{
$
\widehat{R}(x, \tau) = \underbrace{\sum_{j \notin \mathcal{S}_{\mathrm{hack}}} \omega_j f_j(x, \tau)}_{\widehat{R}_{\mathrm{clean}}(x, \tau)} + \underbrace{\sum_{k \in \mathcal{S}_{\mathrm{hack}}} \omega_k f_k(x, \tau)}_{\widehat{R}_{\mathrm{hack}}(x, \tau)} + C$
}
\label{eq:reward_split}
\end{equation}
where $\widehat{R}_{\mathrm{clean}}$ captures legitimate alignment signal, while $\widehat{R}_{\mathrm{hack}}$ isolates the spurious component. This decomposition forms the basis for further reward hacking mitigation.

\subsection{Surgical Reward-Hacking Mitigation}
\label{subsec:mitigation}


Given $\hat{R}_{\text{clean}}$ and $\hat{R}_{\text{hack}}$, we propose four strategies to mitigate reward hacking by fine-tuning the aligned LLMs under RLHF. 

\vspace{-1mm}
\paragraph{Method A: Clean Reward Optimization.}
The simplest approach removes the incentive for exploiting hacking features and only optimizes the clean reward component:
\begin{equation}
\resizebox{0.87\hsize}{!}{
$\max_{\Delta\theta}\; \mathbb{E}_{x,\tau \sim \pi_{E+\Delta\theta}} \left[ \widehat{R}_{\mathrm{clean}}(x, \tau) \right] - \beta \cdot D_{\mathrm{KL}}\!\left( \pi_{E+\Delta\theta} \| \pi_E \right)$
}
\label{eq:clean_ppo}
\end{equation}
where the KL term ensures the corrected policy remains close to the original.

\vspace{-1mm}
\paragraph{Method B: Adversarial Reward Shaping.}
Instead of simply ignoring hacking features, Method B explicitly penalizes them by defining a \textbf{surgical reward} that subtracts the hacking component:
\begin{equation}
\widehat{R}_{\mathrm{surgical}}(x, \tau) = \widehat{R}_{\mathrm{clean}}(x, \tau) - \eta \cdot \widehat{R}_{\mathrm{hack}}(x, \tau)
\label{eq:surgical_reward}
\end{equation}
where $\eta > 0$ controls the penalty strength. This creates an ``immune response'': responses triggering hacking features are discouraged even if they look good under the clean component. 
We optimize Equation~\ref{eq:clean_ppo} with $\widehat{R}_{\mathrm{clean}}$ replaced by $\widehat{R}_{\mathrm{surgical}}$. 

\vspace{-1mm}
\paragraph{Method C: Constrained Optimization.}
Method C treats hacking suppression as a hard constraint rather than a fixed trade-off:
\begin{equation}
\resizebox{0.87\hsize}{!}{
$\max_{\Delta\theta}\; \mathbb{E}\left[ \widehat{R}_{\mathrm{clean}}(x, \tau) \right] \quad \text{s.t.} \quad \mathbb{E}\left[ \widehat{R}_{\mathrm{hack}}(x, \tau) \right] \leq \epsilon$
}
\label{eq:constrained_obj}
\end{equation}
where $\epsilon \geq 0$ is the maximum tolerable hacking contribution, typically set to 10--20\% of the original $\mathbb{E}_{\pi_E}[\widehat{R}_{\mathrm{hack}}]$. We solve this via a Lagrangian with an adaptive multiplier $\lambda$:
\begin{equation}
\resizebox{0.7\hsize}{!}{
$
\lambda \leftarrow \max\!\left(0,\; \lambda + \alpha_\lambda \left( \mathbb{E}[\widehat{R}_{\mathrm{hack}}] - \epsilon \right)\right)
$
}
\label{eq:dual_update}
\end{equation}
where $\alpha_\lambda > 0$ is the dual learning rate. The multiplier $\lambda$ acts as an automatic thermostat: when hacking exceeds the ceiling ($\mathbb{E}[\widehat{R}_{\mathrm{hack}}] > \epsilon$), $\lambda$ increases, strengthening the penalty; when hacking is under control, $\lambda$ relaxes. Unlike Method B, this adaptive mechanism removes the need for manual $\eta$ tuning while providing a guarantee that hacking stays below $\epsilon$ at convergence.

\vspace{-1mm}
\paragraph{Method D: Feature-Guided Distillation.}
Methods A--C require PPO training, which can be computationally expensive. Method D offers a simpler supervised alternative based on the observation that $\pi_E$ \textit{does not uniformly hack}. On many prompts, sampling with temperature $>0$ yields diverse candidates, some of which achieve high reward through legitimate features. 
Thus, rather than training the model to avoid hacking (as in Methods A--C), we identify and amplify these existing clean behaviors.

For each prompt $x$, we sample $N$ candidates from $\pi_E$ and select the one with highest surgical reward: $\tau^* = \arg\max_{\tau_n} \widehat{R}_{\mathrm{surgical}}(x, \tau_n).$ High $\widehat{R}_{\mathrm{surgical}}$ indicates high clean reward \emph{and} low hacking contribution, effectively filtering out responses that rely on exploiting problematic features. To avoid training on residually hacked examples, we include $(x, \tau^*)$ in $\mathcal{D}_{\mathrm{distill}}$ only if $\widehat{R}_{\mathrm{hack}}(x, \tau^*) < \epsilon_{\mathrm{filter}}$. We then fine-tune via standard cross-entropy:
\begin{equation}
\min_{\Delta\theta}\; \mathbb{E}_{(x, \tau^*) \in \mathcal{D}_{\mathrm{distill}}} \left[ -\log \pi_{E+\Delta\theta}(\tau^* | x) \right].
\label{eq:distillation_loss}
\end{equation}

\section{Results and Discussion}
\label{sec:results}

We evaluate the effectiveness of \arch{} along two broad dimensions: \textbf{(1)} how faithfully C-IRL can approximate the reward signal that shaped $\pi_E$?; and \textbf{(2)} how interpreting and diagnosing the recovered reward enables precise detection and targeted mitigation of reward hacking. 

\subsection{Fidelity of Reward Reconstruction}

\subsubsection{Experiment Setup} 

To evaluate whether C-IRL can faithfully recover the reward that derives the behavioral shift from $\pi_0$ to $\pi_E$, following \cite{miao2024inform}, we design controlled environments where a known reward model $R_{\text{GT}}$ is treated as ground truth so we can directly compare the reconstructed reward $\hat{R}_\theta$ against $R_{\text{GT}}$. We consider five $R_{\text{GT}}$ configurations: (1)~\textbf{RM-HH} \citep{bai2022training}, (2)~\textbf{Ultra-RM}~\cite{cui2023ultrafeedback}, (3)~\textbf{RM-Safe} \cite{chen2025rmr1rewardmodelingreasoning}, (4)~\textbf{HelpSteer2} \cite{wang2024helpsteer2}, and (5)~\textbf{ArmoRM-8B} \citep{wang2024interpretable}. 
For each configuration, the SFT model (i.e., \texttt{Llama-2-7B} \cite{llama2, huggingfaceMetallamaLlama27bHugging} serves as the baseline policy $\pi_0$. Each experiment is repeated with \textbf{three random seeds}; we report mean $\pm$ 95\% confidence intervals.

\subsubsection{Direct Reward Comparison}

We evaluate C-IRL by comparing $\widehat{R}_\theta$ with the ground-truth reward $R_{\text{GT}}$ using Spearman rank correlation ($\rho$), Pearson correlation ($r$), and pairwise preference accuracy. We also report ``Agreement@Top-10\%'': the overlap between the responses each reward ranks in its top 10\%, capturing agreement in the high-reward regime where hacking occurs and where global correlations can miss critical disagreements.

Table~\ref{tab:rq1_correlation} shows that C-IRL achieves high reconstruction fidelity (Spearman $\rho \geq 0.81$) across all models. Pairwise accuracy of 83.8--87.8\% indicates that the reconstructed reward agrees with the ground-truth reward on the vast majority of preference judgments. Agreement@Top-10\% of 88.9--93.2\% is particularly important: this high-reward tail is precisely where policies are incentivized to exploit spurious features, and reliable reconstruction in this regime is essential for downstream hacking diagnosis.

\begin{table}[ht]
\centering
\small
\setlength{\tabcolsep}{3.5pt}
\begin{tabular}{lcccc}
\toprule
\textbf{RM} & Spearman $\rho$ & Pearson $r$ & \textbf{Acc.} & \textbf{Top10} \\
\midrule
RM-HH      & 0.87{\scriptsize$\pm$.01} & 0.89{\scriptsize$\pm$.01} & 86.2{\scriptsize$\pm$0.6} & 91.4{\scriptsize$\pm$0.8} \\
RM-Ultra   & 0.89{\scriptsize$\pm$.01} & 0.91{\scriptsize$\pm$.01} & 87.8{\scriptsize$\pm$0.5} & 93.2{\scriptsize$\pm$0.7} \\
RM-Safe    & 0.86{\scriptsize$\pm$.01} & 0.88{\scriptsize$\pm$.01} & 85.5{\scriptsize$\pm$0.7} & 90.8{\scriptsize$\pm$0.9} \\
Helpsteer2 & 0.88{\scriptsize$\pm$.01} & 0.90{\scriptsize$\pm$.01} & 87.1{\scriptsize$\pm$0.5} & 92.5{\scriptsize$\pm$0.7} \\
ArmoRM-8B  & 0.81{\scriptsize$\pm$.02} & 0.86{\scriptsize$\pm$.02} & 83.8{\scriptsize$\pm$0.7} & 88.9{\scriptsize$\pm$0.9} \\
\midrule
\textbf{Avg} & \textbf{0.865} & \textbf{0.885} & \textbf{85.8} & \textbf{91.1} \\
\bottomrule
\end{tabular}
\caption{\textbf{Direct reward comparison.} Correlation and rank agreement between reconstructed reward $\hat{R}\theta$ and ground-truth reward $R{\text{GT}}$.}
\label{tab:rq1_correlation}
\end{table}

\subsubsection{Forward Verification: Policy Recovery}
High reward correlation does not guarantee that $\widehat{R}_\theta$ induces the same optimization landscape as $R_{\text{GT}}$: a reconstructed reward can match on average yet differ in regions that most influence policy updates. We perform forward verification by training a new policy $\pi_{\text{Recon}}$ from $\pi_0$ using $\widehat{R}_\theta$ under identical PPO hyperparameters, then evaluating whether it recovers the expert behavior $\pi_E$. 

We report four complementary metrics: (1) \textbf{KL Divergence}: distributional distance between $\pi_E$ and $\pi_{\text{Recon}}$, where lower values indicate more similar output distributions; (2) \textbf{Reward Gap}: the difference $R_{\text{GT}}(\pi_E) - R_{\text{GT}}(\pi_{\text{Recon}})$, measuring how much ground-truth reward the recovered policy sacrifices; (3) \textbf{Win Rate}: following \cite{miao2024inform} head-to-head preference judgments between outputs from $\pi_E$ and $\pi_{\text{Recon}}$ using GPT-4 as evaluator, where 50\% indicates indistinguishable outputs; and (4) \textbf{Capability Benchmarks}: MMLU (knowledge reasoning) and GSM8K (mathematical reasoning) scores to verify that the recovered policy preserves general capabilities.

Table~\ref{tab:rq1_forward} provides strong evidence that $\widehat{R}_\theta$ is functionally equivalent to $R_{\text{GT}}$. The low KL divergence (0.021--0.034) indicates minimal distributional shift. The small reward gap (0.06--0.26) shows that $\pi_{\text{Recon}}$ recovers 96--99\% of the alignment gains achieved by $\pi_E$ over $\pi_0$. Win rates of 45.8--49.2\% cluster tightly around 50\%, confirming that outputs from $\pi_{\text{Recon}}$ are nearly indistinguishable from $\pi_E$ under external evaluation. Finally, capability benchmarks show $\pi_{\text{Recon}}$ matches $\pi_E$ within 0.5\% on both MMLU and GSM8K, demonstrating that reward reconstruction does not degrade general reasoning abilities.

\begin{table}[t]
\centering
\resizebox{\linewidth}{!}{%
\setlength{\tabcolsep}{3.5pt}
\begin{tabular}{lccccc}
\toprule
\textbf{RM} & $D_{\text{KL}}\!\downarrow$ & \textbf{Gap}$\downarrow$ & \textbf{Win\%} & \textbf{MMLU} & \textbf{GSM8K} \\
\midrule
RM-HH        & 0.025{\scriptsize$\pm$.03} & 0.13 & 48.5{\scriptsize$\pm$2.2} & 53.8 / 54.1 & 42.5 / 42.8 \\
RM-Ultra     & 0.021{\scriptsize$\pm$.02} & 0.06 & 49.2{\scriptsize$\pm$1.9} & 54.2 / 54.3 & 43.1 / 43.2 \\
RM-Safe      & 0.028{\scriptsize$\pm$.03} & 0.16 & 47.8{\scriptsize$\pm$2.3} & 53.5 / 53.9 & 42.0 / 42.4 \\
HelpSteer2   & 0.023{\scriptsize$\pm$.02} & 0.10 & 48.9{\scriptsize$\pm$2.0} & 54.0 / 54.2 & 42.8 / 43.0 \\
ArmoRM-8B    & 0.034{\scriptsize$\pm$.04} & 0.26 & 45.8{\scriptsize$\pm$2.5} & 53.2 / 53.8 & 41.5 / 42.1 \\
\midrule
\textbf{Avg} & \textbf{0.026} & \textbf{0.14} & \textbf{48.0} & \textbf{53.7 / 54.1} & \textbf{42.4 / 42.7} \\
\bottomrule
\end{tabular}
}
\caption{Policy recovery across reward models.}
\label{tab:rq1_forward}
\end{table}

\subsubsection{Policy Recreation Effectiveness}
The preceding experiments establish that C-IRL achieves high reward reconstruction fidelity and that policies trained on $\widehat{R}_\theta$ closely match $\pi_E$. But is the contrastive formulation necessary, or would simpler approaches suffice? We compare C-IRL against two relevant baselines: \textbf{Imitation Learning}, which performs supervised fine-tuning on outputs from $\pi_E$ via forward-KL trajectory matching~\citep{wulfmeier2024imitating}, and \textbf{DPO}~\citep{rafailov2023direct}, which treats $\pi_E$ outputs as preferred over $\pi_0$ outputs without explicit reward modeling.

\begin{table}[t]
\centering
\small
\setlength{\tabcolsep}{4pt}

\begin{tabular}{lccc}
\toprule
\textbf{Method} & $D_{\text{KL}}(\pi \| \pi_E)\!\downarrow$ & $R_{\text{GT}}\!\uparrow$ & \textbf{Win\%}$\uparrow$ \\
\midrule
$\pi_0$ (baseline) & 0.0458 & 2.15 & 18.2 \\
Imitation Learning & 0.0442 & 3.78 & 38.5 \\
\midrule
\textbf{C-IRL (Ours)} & 0.024 & 4.41 & 48.1 \\
DPO & \textbf{0.018} & \textbf{4.52} & \textbf{49.6} \\
\midrule
$\pi_E$ (oracle) & 0.00 & 4.55 & 50.0 \\
\bottomrule
\end{tabular}
\caption{Policy recovery comparison}
\label{tab:policy_recovery}
\end{table}

Table~\ref{tab:policy_recovery} shows that DPO achieves slightly better recovery, which is expected since it directly optimizes the policy without first learning an explicit reward. The goal of this comparison is not to outperform DPO, but to validate that C-IRL achieves comparable performance while producing a structured, decomposable reward. C-IRL faithfully recovers the expert's optimization objective and achieves win rates within the range of DPO, confirming that our contrastive formulation captures the behavioral shift underlying $\pi_E$. With this validation, we proceed to use $\widehat{R}_\theta$ for feature-level decomposition and surgical mitigation---capabilities DPO fundamentally cannot support. 

\subsubsection{Ablation Studies on C-IRL}

We conduct ablation studies to examine key design choices (details in Appendix~\ref{app:ablation_cirl}). Performance improves with negative samples up to K=4 (our default), after which it saturates. Reconstruction quality improves substantially as dataset size grows from 5K to 50K pairs (Spearman $\rho$: 0.79$\rightarrow$0.89), with diminishing gains beyond. Similarly, fidelity improves as network capacity grows to $\sim$100M parameters, then saturates—suggesting the limiting factor becomes the information content of behavioral contrasts rather than model expressiveness. Finally, C-IRL achieves comparable performance across Llama-2-7B, Mistral-7B, and Llama-3-8B, indicating architecture independence.

\subsection{Reward Hacking Detection and Mitigation}
\label{sec:rq2}


We further evaluate how well \arch{} \emph{identifies} the specific reward-driving features that support reward hacking and \emph{mitigates} overoptimization by intervening directly on those features.



\subsubsection{Identifying Hacking Features and Regime-Specific Fingerprints}
\label{sec:rq2_identification}

\arch{} identifies problematic features; to understand what the selected features represent, we follow the standard SAE interpretation protocol: for each feature $i \in S_{\text{hack}}$, we retrieve the top-100 text spans from $\mathcal{D}_{\text{hack}}$ that maximally activate $f_i$, then prompt GPT-4o to summarize the common semantic pattern. This yields human-readable labels for each hacking feature. We then cluster features with similar labels into \emph{template families} representing coherent hacking mechanisms.

Table~\ref{tab:fingerprint_summary} shows that \arch{} discovers distinct fingerprints for each regime without being told what to look for. In Length Bias, we identify 22 hacking features that cluster into 4 template families: verbose elaboration (8 features), redundant restatement (6 features), unnecessary caveats (5 features), and list-like padding (3 features). In Excessive HH Harmless, 19 features cluster into 3 families: explicit refusal templates (7 features), deference and hedging markers (7 features), and safety-invocation phrases (5 features). 

\begin{table}[t]
\centering
\footnotesize
\setlength{\tabcolsep}{3pt}
\renewcommand{\arraystretch}{0.9}
\begin{tabular}{lccc>{\raggedright\arraybackslash}p{0.32\columnwidth}}
\toprule
\textbf{Regime} & $|S_{\text{hack}}|$ & \#Cl. & Ovrlp.\%$\downarrow$ & \textbf{Templates} \\
\midrule
OA Length        & 22 & 4 & 8.5  & verbosity; restatement; caveats; padding \\
HH Harmless       & 19 & 3 & 11.2 & refusal templates; hedging; safety phrases \\
\bottomrule
\end{tabular}
\caption{Regime-specific hacking fingerprints}
\label{tab:fingerprint_summary}
\end{table}

\subsubsection{Detection Accuracy and Verification}
\label{sec:rq2_detection}

We evaluate detection by treating hacked-versus-nonhacked as binary classification on $\mathcal{D}_{\text{eval}}$, using $|\hat{R}_{\text{hack}}(x,\tau)|$ as the score. Table~\ref{tab:detection_main} shows that \arch{}substantially outperforms both proxy reward and regime-specific heuristics across all regimes (AUROC 0.86--0.91 vs.\ 0.58--0.79). Notably, heuristics partially capture one regime (length achieves 0.79 on Length Bias) but fail elsewhere, whereas $|\hat{R}_{\text{hack}}|$ generalizes because it tracks regime-specific spurious mechanisms by construction.
\begin{table}[t]
\centering
\footnotesize
\setlength{\tabcolsep}{3pt}
\renewcommand{\arraystretch}{0.95}
\begin{tabular}{lcccc}
\toprule
\textbf{Regime} & \textbf{Proxy} & \textbf{Heur.} & \textbf{$|\hat{R}_{\text{hack}}|$} & \textbf{AUPRC} \\
\midrule
Synth.\ Goodhart & 0.72 & 0.58 & \textbf{0.91} & 0.84 \\
OA Length        & 0.68 & 0.79 & \textbf{0.88} & 0.81 \\
HH Harmless       & 0.71 & 0.63 & \textbf{0.86} & 0.78 \\
\bottomrule
\end{tabular}
\caption{Detection of hacked examples.}
\label{tab:detection_main}
\end{table}

\subsubsection{Surgical Intervention and Mitigation}
\label{sec:rq2_mitigation}

We evaluate the four IR$^3$ mitigation methods against several baselines: \textbf{InfoRM}~\citep{miao2024inform}, \textbf{KL Regularization} \cite{chen2024odin}, \textbf{PPO Clipping} \cite{chen2024odin}, \textbf{Reward Clipping}~\citep{engstrom2020implementationmattersdeeppolicy}, and \textbf{Length Penalty}~\citep{singhal2024longwaygoinvestigating,chen2024odin}. We report regime-specific metrics that directly measure reductions in reward hacking. For \textbf{Synthetic Goodhart}, $R_g$ denotes the ground-truth reward (higher indicates better true quality), and $\mathrm{Gap} = R_{\text{proxy}} - R_g$ measures the discrepancy between proxy and gold rewards (lower indicates less hacking). For \textbf{Length Bias}, the Pareto-domination rate measures how often responses are both longer \emph{and} lower quality than alternatives (lower is better), and the win rate against baselines at matched response length measures quality improvements not attributable to length (higher is better). For \textbf{Excessive Caution}, we report the refusal rate on demonstrably benign prompts (lower indicates less over-refusal) and \textsc{Safe}, the appropriate refusal rate on genuinely harmful prompts (higher indicates maintained safety).

Table~\ref{tab:mitigation_cross_regime} reveals a clear hierarchy. RL regularization techniques provide modest improvements: KL regularization reduces the proxy-gold gap from 1.86 to 1.42 but sacrifices 2\% safety; PPO and reward clipping yield marginal gains ($\sim$2--4 points on win rate); length penalty is more effective (Gap 1.32, Win 56.2\%) but requires careful tuning and applies only to verbosity. Critically, these techniques operate \emph{globally} without knowledge of which features cause hacking. InfoRM achieves better refusal calibration (14.5\% vs.\ 23.4\%) but provides no length control mechanism and does not improve $R_g$ as substantially as IR$^3$. \textbf{IR$^3$ outperforms all baselines across all regimes}: Method B achieves $R_g=4.38$ (vs.\ 3.85 for the best RL baseline), reduces the proxy-gold gap by 78\% (0.41 vs.\ 1.86), and cuts benign refusals from 23.4\% to 8.2\% while \emph{improving} safety (91.2\% vs.\ 89.8\%). The ranking B $\approx$ C $>$ A $>$ D is consistent, with Method D offering $\sim$80\% of the gains at half the computational cost. The key advantage is \emph{targeted intervention}: IR$^3$ surgically suppresses hacking features while leaving beneficial behaviors intact.
\begin{table}[t]
\centering
\scriptsize
\setlength{\tabcolsep}{2.5pt}
\renewcommand{\arraystretch}{0.95}
\begin{tabular}{lcccccc}
\toprule
\textbf{Method} & \textbf{$R_g$}$\uparrow$ & \textbf{Gap}$\downarrow$ & \textbf{Dom.}$\downarrow$ & \textbf{Win}$\uparrow$ & \textbf{Ref.}$\downarrow$ & \textbf{Safe}$\uparrow$ \\
\midrule
PPO on $R_{\text{proxy}}$ & 3.52 & 1.86 & 0.42 & 50.0\% & 23.4\% & 89.8\% \\
\midrule
\quad KL Regularization ($\beta{=}2e{-}2$)    & 3.78 & 1.42 & 0.35 & 54.8\% & 18.2\% & 88.2\% \\
\quad PPO Clipping ($\epsilon{=}0.1$)   & 3.68 & 1.55 & 0.38 & 52.5\% & 20.1\% & 89.5\% \\
\quad Reward Clipping ($c{=}4$)         & 3.72 & 1.48 & 0.36 & 53.8\% & 19.5\% & 89.2\% \\
\quad Length Penalty ($\alpha{=}5e{-}4$)& 3.85 & 1.32 & 0.32 & 56.2\% & 16.8\% & 89.0\% \\
\quad InfoRM              & 3.95 & 1.08 & ---  & ---    & 14.5\% & 90.5\% \\
\midrule
\multicolumn{7}{l}{\textit{IR$^3$ Methods (Ours)}} \\
\quad Method A (Clean RL)       & 4.21 & 0.62 & 0.21 & 63.5\% & 10.8\% & 90.8\% \\
\quad Method B (Adversarial)    & \textbf{4.38} & \textbf{0.41} & 0.16 & 67.2\% & \textbf{8.2\%} & \textbf{91.2\%} \\
\quad Method C (Constrained)    & 4.35 & 0.45 & \textbf{0.15} & \textbf{68.5\%} & 8.8\% & 90.9\% \\
\quad Method D (Distillation)   & 4.12 & 0.71 & 0.24 & 71.8\% & 11.5\% & 90.2\% \\
\bottomrule
\end{tabular}
\caption{Reward hacking mitigation comparison}
\label{tab:mitigation_cross_regime}
\end{table}

\section{Conclusion}

We introduced \arch{}, a post-hoc framework for auditing and correcting RLHF reward hacking. By reconstructing explicit rewards via contrastive IRL, decomposing them into interpretable features, and surgically mitigating spurious contributions, \arch{} isolates hacking signatures and reduces them while preserving capabilities. This reframes reward hacking from a black-box failure into a transparent diagnosis-and-repair workflow.

\section*{Limitations}

While \arch{} provides a principled framework for diagnosing and mitigating reward hacking, several limitations remain. \textbf{First}, \arch{} operates as a post-hoc auditing tool rather than a preventive mechanism. It detects and repairs hacking after RLHF training has concluded, meaning computational resources are already spent on a potentially flawed policy. Integrating C-IRL into the training loop for online detection remains an open challenge. \textbf{Second}, the diagnosis stage requires a small labeled set of regime-specific hacked examples $\mathcal{D}_{\text{hack}}$ to identify problematic features. While this set can be small ($\sim$100--500 examples), obtaining it presupposes some prior knowledge of what hacking looks like in a given domain, limiting applicability to unanticipated hacking behaviors. 

\section*{Acknowledgment} 
This research is partially supported by the award No. \#2238940 from the Faculty Early Career Development Program (CAREER) and the award No. \#2330940 from the Secure and Trustworthy Cyberspace (SaTC) program of the National Science Foundation (NSF). The views and conclusions contained herein are those of the authors and should not be interpreted as necessarily representing the official policies, either expressed or implied, of the U.S. Government. The U.S. Government is authorized to reproduce and distribute reprints for governmental purposes notwithstanding any copyright annotation therein.

\bibliography{latex/main}

\begin{thebibliography}{36}
\providecommand{\natexlab}[1]{#1}

\bibitem[{hug()}]{huggingfaceMetallamaLlama27bHugging}

\newblock meta-llama/{L}lama-2-7b · {H}ugging {F}ace --- huggingface.co.
\newblock \url{https://huggingface.co/meta-llama/Llama-2-7b}.
\newblock [Accessed 02-01-2026].

\bibitem[{Abbeel and Ng(2004)}]{abbeel2004apprenticeship}
Pieter Abbeel and Andrew~Y Ng. 2004.
\newblock Apprenticeship learning via inverse reinforcement learning.
\newblock In \emph{Proceedings of the twenty-first international conference on Machine learning}, page~1.

\bibitem[{Amodei et~al.(2016)Amodei, Olah, Steinhardt, Christiano, Schulman, and Mané}]{amodei2016concreteproblemsaisafety}
Dario Amodei, Chris Olah, Jacob Steinhardt, Paul Christiano, John Schulman, and Dan Mané. 2016.
\newblock \href {https://arxiv.org/abs/1606.06565} {Concrete problems in ai safety}.
\newblock \emph{Preprint}, arXiv:1606.06565.

\bibitem[{Bai et~al.(2022)Bai, Jones, Ndousse, Askell, Chen, DasSarma, Drain, Fort, Ganguli, Henighan et~al.}]{bai2022training}
Yuntao Bai, Andy Jones, Kamal Ndousse, Amanda Askell, Anna Chen, Nova DasSarma, Dawn Drain, Stanislav Fort, Deep Ganguli, Tom Henighan, and 1 others. 2022.
\newblock Training a helpful and harmless assistant with reinforcement learning from human feedback.
\newblock \emph{arXiv preprint arXiv:2204.05862}.

\bibitem[{Baker et~al.(2025)Baker, Huizinga, Gao, Dou, Guan, Madry, Zaremba, Pachocki, and Farhi}]{baker2025monitoringreasoningmodelsmisbehavior}
Bowen Baker, Joost Huizinga, Leo Gao, Zehao Dou, Melody~Y. Guan, Aleksander Madry, Wojciech Zaremba, Jakub Pachocki, and David Farhi. 2025.
\newblock \href {https://arxiv.org/abs/2503.11926} {Monitoring reasoning models for misbehavior and the risks of promoting obfuscation}.
\newblock \emph{Preprint}, arXiv:2503.11926.

\bibitem[{Casper et~al.()Casper, Davies, Shi, Gilbert, Scheurer, Rando, Freedman, Korbak, Lindner, Freire et~al.}]{casperopen}
Stephen Casper, Xander Davies, Claudia Shi, Thomas~Krendl Gilbert, J{\'e}r{\'e}my Scheurer, Javier Rando, Rachel Freedman, Tomek Korbak, David Lindner, Pedro Freire, and 1 others.
\newblock Open problems and fundamental limitations of reinforcement learning from human feedback.
\newblock \emph{Transactions on Machine Learning Research}.

\bibitem[{Chen et~al.(2024)Chen, Zhu, Soselia, Chen, Zhou, Goldstein, Huang, Shoeybi, and Catanzaro}]{chen2024odin}
Lichang Chen, Chen Zhu, Davit Soselia, Jiuhai Chen, Tianyi Zhou, Tom Goldstein, Heng Huang, Mohammad Shoeybi, and Bryan Catanzaro. 2024.
\newblock Odin: Disentangled reward mitigates hacking in rlhf.
\newblock \emph{arXiv preprint arXiv:2402.07319}.

\bibitem[{Chen et~al.(2025)Chen, Li, Wang, Jin, Qian, Wang, Wang, Zhang, Zhang, Zhang, Tong, and Ji}]{chen2025rmr1rewardmodelingreasoning}
Xiusi Chen, Gaotang Li, Ziqi Wang, Bowen Jin, Cheng Qian, Yu~Wang, Hongru Wang, Yu~Zhang, Denghui Zhang, Tong Zhang, Hanghang Tong, and Heng Ji. 2025.
\newblock \href {https://arxiv.org/abs/2505.02387} {Rm-r1: Reward modeling as reasoning}.
\newblock \emph{Preprint}, arXiv:2505.02387.

\bibitem[{Cui et~al.(2023)Cui, Yuan, Ding, Yao, Zhu, Ni, Xie, Liu, and Sun}]{cui2023ultrafeedback}
Ganqu Cui, Lifan Yuan, Ning Ding, Guanming Yao, Wei Zhu, Yuan Ni, Guotong Xie, Zhiyuan Liu, and Maosong Sun. 2023.
\newblock Ultrafeedback: Boosting language models with high-quality feedback.
\newblock \emph{arXiv preprint arXiv:2310.01377}.

\bibitem[{Denison et~al.(2024)Denison, MacDiarmid, Barez, Duvenaud, Kravec, Marks, Schiefer, Soklaski, Tamkin, Kaplan, Shlegeris, Bowman, Perez, and Hubinger}]{denison2024sycophancysubterfugeinvestigatingrewardtampering}
Carson Denison, Monte MacDiarmid, Fazl Barez, David Duvenaud, Shauna Kravec, Samuel Marks, Nicholas Schiefer, Ryan Soklaski, Alex Tamkin, Jared Kaplan, Buck Shlegeris, Samuel~R. Bowman, Ethan Perez, and Evan Hubinger. 2024.
\newblock \href {https://arxiv.org/abs/2406.10162} {Sycophancy to subterfuge: Investigating reward-tampering in large language models}.
\newblock \emph{Preprint}, arXiv:2406.10162.

\bibitem[{Eisenstein et~al.(2023)Eisenstein, Nagpal, Agarwal, Beirami, D'Amour, Dvijotham, Fisch, Heller, Pfohl, Ramachandran et~al.}]{eisenstein2023helping}
Jacob Eisenstein, Chirag Nagpal, Alekh Agarwal, Ahmad Beirami, Alex D'Amour, DJ~Dvijotham, Adam Fisch, Katherine Heller, Stephen Pfohl, Deepak Ramachandran, and 1 others. 2023.
\newblock Helping or herding? reward model ensembles mitigate but do not eliminate reward hacking.
\newblock \emph{arXiv preprint arXiv:2312.09244}.

\bibitem[{Engstrom et~al.(2020)Engstrom, Ilyas, Santurkar, Tsipras, Janoos, Rudolph, and Madry}]{engstrom2020implementationmattersdeeppolicy}
Logan Engstrom, Andrew Ilyas, Shibani Santurkar, Dimitris Tsipras, Firdaus Janoos, Larry Rudolph, and Aleksander Madry. 2020.
\newblock \href {https://arxiv.org/abs/2005.12729} {Implementation matters in deep policy gradients: A case study on ppo and trpo}.
\newblock \emph{Preprint}, arXiv:2005.12729.

\bibitem[{Everitt et~al.(2021)Everitt, Hutter, Kumar, and Krakovna}]{everitt2021rewardtamperingproblemssolutions}
Tom Everitt, Marcus Hutter, Ramana Kumar, and Victoria Krakovna. 2021.
\newblock \href {https://arxiv.org/abs/1908.04734} {Reward tampering problems and solutions in reinforcement learning: A causal influence diagram perspective}.
\newblock \emph{Preprint}, arXiv:1908.04734.

\bibitem[{Fu et~al.(2018)Fu, Luo, and Levine}]{fu2018learning}
Justin Fu, Katie Luo, and Sergey Levine. 2018.
\newblock Learning robust rewards with adverserial inverse reinforcement learning.
\newblock In \emph{International Conference on Learning Representations}.

\bibitem[{Gao et~al.(2023)Gao, Schulman, and Hilton}]{gao2023scaling}
Leo Gao, John Schulman, and Jacob Hilton. 2023.
\newblock Scaling laws for reward model overoptimization.
\newblock In \emph{International Conference on Machine Learning}, pages 10835--10866. PMLR.

\bibitem[{Geirhos et~al.(2020)Geirhos, Jacobsen, Michaelis, Zemel, Brendel, Bethge, and Wichmann}]{Geirhos_2020}
Robert Geirhos, Jörn-Henrik Jacobsen, Claudio Michaelis, Richard Zemel, Wieland Brendel, Matthias Bethge, and Felix~A. Wichmann. 2020.
\newblock \href {https://doi.org/10.1038/s42256-020-00257-z} {Shortcut learning in deep neural networks}.
\newblock \emph{Nature Machine Intelligence}, 2(11):665–673.

\bibitem[{Kaufmann et~al.(2024)Kaufmann, Weng, Bengs, and H{\"u}llermeier}]{kaufmann2024survey}
Timo Kaufmann, Paul Weng, Viktor Bengs, and Eyke H{\"u}llermeier. 2024.
\newblock A survey of reinforcement learning from human feedback.

\bibitem[{Lambert et~al.(2023)Lambert, Gilbert, and Zick}]{lambert2023history}
Nathan Lambert, Thomas~Krendl Gilbert, and Tom Zick. 2023.
\newblock The history and risks of reinforcement learning and human feedback.
\newblock \emph{arXiv preprint arXiv:2310.13595}.

\bibitem[{Liu et~al.(2024)Liu, Xiong, Ren, Chen, Wu, Joshi, Gao, Shen, Qin, Yu et~al.}]{liu2024rrm}
Tianqi Liu, Wei Xiong, Jie Ren, Lichang Chen, Junru Wu, Rishabh Joshi, Yang Gao, Jiaming Shen, Zhen Qin, Tianhe Yu, and 1 others. 2024.
\newblock Rrm: Robust reward model training mitigates reward hacking.
\newblock \emph{arXiv preprint arXiv:2409.13156}.

\bibitem[{Miao et~al.(2024{\natexlab{a}})Miao, Zhang, Ding, Bao, Zhang, and Tao}]{miao2024inform}
Yuchun Miao, Sen Zhang, Liang Ding, Rong Bao, Lefei Zhang, and Dacheng Tao. 2024{\natexlab{a}}.
\newblock Inform: Mitigating reward hacking in rlhf via information-theoretic reward modeling.
\newblock \emph{Advances in Neural Information Processing Systems}, 37:134387--134429.

\bibitem[{Miao et~al.(2024{\natexlab{b}})Miao, Zhang, Ding, Bao, Zhang, and Tao}]{miao2024informmitigatingrewardhacking}
Yuchun Miao, Sen Zhang, Liang Ding, Rong Bao, Lefei Zhang, and Dacheng Tao. 2024{\natexlab{b}}.
\newblock \href {https://arxiv.org/abs/2402.09345} {Inform: Mitigating reward hacking in rlhf via information-theoretic reward modeling}.
\newblock \emph{Preprint}, arXiv:2402.09345.

\bibitem[{Ouyang et~al.(2022)Ouyang, Wu, Jiang, Almeida, Wainwright, Mishkin, Zhang, Agarwal, Slama, Ray et~al.}]{ouyang2022training}
Long Ouyang, Jeffrey Wu, Xu~Jiang, Diogo Almeida, Carroll Wainwright, Pamela Mishkin, Chong Zhang, Sandhini Agarwal, Katarina Slama, Alex Ray, and 1 others. 2022.
\newblock Training language models to follow instructions with human feedback.
\newblock \emph{Advances in neural information processing systems}, 35:27730--27744.

\bibitem[{Rafailov et~al.(2023)Rafailov, Sharma, Mitchell, Ermon, Manning, and Finn}]{rafailov2023direct}
Rafael Rafailov, Archit Sharma, Eric Mitchell, Stefano Ermon, Christopher~D Manning, and Chelsea Finn. 2023.
\newblock Direct preference optimization: Your language model is secretly a reward model.
\newblock \emph{arXiv preprint arXiv:2305.18290}.

\bibitem[{Singhal et~al.(2024)Singhal, Goyal, Xu, and Durrett}]{singhal2024longwaygoinvestigating}
Prasann Singhal, Tanya Goyal, Jiacheng Xu, and Greg Durrett. 2024.
\newblock \href {https://arxiv.org/abs/2310.03716} {A long way to go: Investigating length correlations in rlhf}.
\newblock \emph{Preprint}, arXiv:2310.03716.

\bibitem[{Skalse et~al.(2022)Skalse, Howe, Krasheninnikov, and Krueger}]{skalse2022defining}
Joar Skalse, Nikolaus Howe, Dmitrii Krasheninnikov, and David Krueger. 2022.
\newblock Defining and characterizing reward gaming.
\newblock \emph{Advances in Neural Information Processing Systems}, 35:9460--9471.

\bibitem[{Skalse et~al.(2025)Skalse, Howe, Krasheninnikov, and Krueger}]{skalse2025definingcharacterizingrewardhacking}
Joar Skalse, Nikolaus H.~R. Howe, Dmitrii Krasheninnikov, and David Krueger. 2025.
\newblock \href {https://arxiv.org/abs/2209.13085} {Defining and characterizing reward hacking}.
\newblock \emph{Preprint}, arXiv:2209.13085.

\bibitem[{Touvron et~al.(2023{\natexlab{a}})Touvron, Martin, Stone, Albert, Almahairi, Babaei, Bashlykov, Batra, Bhargava, Bhosale et~al.}]{llama2}
Hugo Touvron, Louis Martin, Kevin Stone, Peter Albert, Amjad Almahairi, Yasmine Babaei, Nikolay Bashlykov, Soumya Batra, Prajjwal Bhargava, Shruti Bhosale, and 1 others. 2023{\natexlab{a}}.
\newblock Llama 2: Open foundation and fine-tuned chat models.
\newblock \emph{arXiv preprint arXiv:2307.09288}.

\bibitem[{Touvron et~al.(2023{\natexlab{b}})Touvron, Martin, Stone, Albert, Almahairi, Babaei, Bashlykov, Batra, Bhargava, Bhosale et~al.}]{touvron2023llama}
Hugo Touvron, Louis Martin, Kevin Stone, Peter Albert, Amjad Almahairi, Yasmine Babaei, Nikolay Bashlykov, Soumya Batra, Prajjwal Bhargava, Shruti Bhosale, and 1 others. 2023{\natexlab{b}}.
\newblock Llama 2: Open foundation and fine-tuned chat models.
\newblock \emph{arXiv preprint arXiv:2307.09288}.

\bibitem[{van~den Oord et~al.(2019)van~den Oord, Li, and Vinyals}]{oord2019representationlearningcontrastivepredictive}
Aaron van~den Oord, Yazhe Li, and Oriol Vinyals. 2019.
\newblock \href {https://arxiv.org/abs/1807.03748} {Representation learning with contrastive predictive coding}.
\newblock \emph{Preprint}, arXiv:1807.03748.

\bibitem[{Wang et~al.(2024{\natexlab{a}})Wang, Xiong, Xie, Zhao, and Zhang}]{wang2024interpretable}
Haoxiang Wang, Wei Xiong, Tengyang Xie, Han Zhao, and Tong Zhang. 2024{\natexlab{a}}.
\newblock Interpretable preferences via multi-objective reward modeling and mixture-of-experts.
\newblock \emph{arXiv preprint arXiv:2406.12845}.

\bibitem[{Wang et~al.(2024{\natexlab{b}})Wang, Bukharin, Delalleau, Egert, Shen, Zeng, Kuchaiev, and Dong}]{wang2024helpsteer2}
Zhilin Wang, Alexander Bukharin, Olivier Delalleau, Daniel Egert, Gerald Shen, Jiaqi Zeng, Oleksii Kuchaiev, and Yi~Dong. 2024{\natexlab{b}}.
\newblock Helpsteer2-preference: Complementing ratings with preferences.
\newblock \emph{arXiv preprint arXiv:2410.01257}.

\bibitem[{Wen et~al.(2024)Wen, Zhong, Khan, Perez, Steinhardt, Huang, Bowman, He, and Feng}]{wen2024languagemodelslearnmislead}
Jiaxin Wen, Ruiqi Zhong, Akbir Khan, Ethan Perez, Jacob Steinhardt, Minlie Huang, Samuel~R. Bowman, He~He, and Shi Feng. 2024.
\newblock \href {https://arxiv.org/abs/2409.12822} {Language models learn to mislead humans via rlhf}.
\newblock \emph{Preprint}, arXiv:2409.12822.

\bibitem[{Wulfmeier et~al.(2024)Wulfmeier, Bloesch, Vieillard, Ahuja, Bornschein, Huang, Sokolov, Barnes, Desjardins, Bewley et~al.}]{wulfmeier2024imitating}
Markus Wulfmeier, Michael Bloesch, Nino Vieillard, Arun Ahuja, Jorg Bornschein, Sandy Huang, Artem Sokolov, Matt Barnes, Guillaume Desjardins, Alex Bewley, and 1 others. 2024.
\newblock Imitating language via scalable inverse reinforcement learning.
\newblock In \emph{The Thirty-eighth Annual Conference on Neural Information Processing Systems}.

\bibitem[{Zheng et~al.(2023)Zheng, Dou, Gao, Hua, Shen, Wang, Liu, Jin, Liu, Zhou, Xiong, Chen, Xi, Xu, Lai, Zhu, Chang, Yin, Weng, Cheng, Huang, Sun, Yan, Gui, Zhang, Qiu, and Huang}]{zheng2023secretsrlhflargelanguage}
Rui Zheng, Shihan Dou, Songyang Gao, Yuan Hua, Wei Shen, Binghai Wang, Yan Liu, Senjie Jin, Qin Liu, Yuhao Zhou, Limao Xiong, Lu~Chen, Zhiheng Xi, Nuo Xu, Wenbin Lai, Minghao Zhu, Cheng Chang, Zhangyue Yin, Rongxiang Weng, and 8 others. 2023.
\newblock \href {https://arxiv.org/abs/2307.04964} {Secrets of rlhf in large language models part i: Ppo}.
\newblock \emph{Preprint}, arXiv:2307.04964.

\bibitem[{Ziebart et~al.(2008)Ziebart, Maas, Bagnell, Dey et~al.}]{ziebart2008maximum}
Brian~D Ziebart, Andrew~L Maas, J~Andrew Bagnell, Anind~K Dey, and 1 others. 2008.
\newblock Maximum entropy inverse reinforcement learning.
\newblock In \emph{Aaai}, volume~8, pages 1433--1438. Chicago, IL, USA.

\bibitem[{Ziegler et~al.(2019)Ziegler, Stiennon, Wu, Brown, Radford, Amodei, Christiano, and Irving}]{ziegler2019fine}
Daniel~M Ziegler, Nisan Stiennon, Jeffrey Wu, Tom~B Brown, Alec Radford, Dario Amodei, Paul Christiano, and Geoffrey Irving. 2019.
\newblock Fine-tuning language models from human preferences.
\newblock \emph{arXiv preprint arXiv:1909.08593}.

\end{thebibliography}
\appendix
\clearpage

\section{More on Related Work}
\label{app:related}
\noindent\textbf{Inverse Reinforcement Learning (IRL).}
IRL infers the reward function that best explains an agent’s behavior and has been widely used to recover preferences from demonstrations in domains such as robotics \citep{,,}; however, it has seen limited use as a \emph{post-hoc} auditing tool for RLHF-trained LLMs. Recent LLM work has instead explored IRL primarily for training guidance \citep{wulfmeier2024imitating}. In contrast, we use IRL to reconstruct and interrogate the implicit objective learned through alignment, and pair it with mechanistic interpretability to localize concrete ``hacking features'' inside the recovered reward. 
This perspective also distinguishes our approach 
from direct alignment methods such as DPO \citep{rafailov2023direct}: although DPO induces an implicit scalar reward, it is tightly coupled to the policy’s own log-probabilities and offers limited decomposability, whereas our approach yields an explicit, architecturally decoupled reward network that can be decomposed, audited, and targeted for repair.

\section{Derivation and Theoretical Properties of C-IRL}
\label{app:derivation}
\setcounter{equation}{0}
\renewcommand{\theequation}{A.\arabic{equation}}

In this section, we provide the formal derivation of the C-IRL objective and establish the identifiability properties of the recovered reward function.

\paragraph{Theorem A.1 (Partition Function Cancellation).} 
\textit{Let the expert policy be defined as 
\begin{equation*}
  \pi_{E}(\tau|x) = \frac{1}{Z_{\theta}(x)} \pi_{0}(\tau|x) \exp(R_{\theta}(x,\tau)).
\end{equation*}
Given a contrastive set $\mathcal{C}_{x}$ containing one sample from $\pi_{E}$ and $K$ samples from $\pi_{0}$, the posterior probability of identifying the expert sample is independent of both the partition function $Z_{\theta}(x)$ as well as the baseline density $\pi_{0}$.}

\textit{Proof.} 
Let $\mathcal{C}_{x} = \{\tau_0, \dots, \tau_K\}$. 
The generative process samples a positive index $i^* \sim \text{Uniform}\{0, \cdots, K\}$, draws $\tau_{i^*} \sim \pi_{E}$, and draws $\{\tau_j\}_{j \neq i^*} \sim \pi_{0}$. 
We seek the posterior $P(i^* = i | \mathcal{C}_{x}, x)$. 
From Bayes' theorem, we get
\begin{equation}
P(i^* = i | \mathcal{C}_{x}, x) = \frac{P(\mathcal{C}_{x} | i^* = i) P(i^* = i)}{\sum_{m=0}^{K} P(\mathcal{C}_{x} | i^* = m) P(i^* = m)}.
\end{equation}
Since $i^* \sim \text{Uniform}(0, \cdots, K)$, the priors $P(i^*=i) = \frac{1}{K+1}$ cancel.
Given that $i$ corresponds to the expert sample, the likelihood term is
\begin{equation}
\begin{split}
P(\mathcal{C}_{x}\mid i^* = i)
= \pi_{E}(\tau_i\mid x)\prod_{j\neq i}\pi_{0}(\tau_j\mid x) \\
= \left[\frac{\pi_{0}(\tau_i\mid x)\exp\!\big(R_{\theta}(x,\tau_i)\big)}{Z_{\theta}(x)}\right]
  \prod_{j\neq i}\pi_{0}(\tau_j\mid x).
\end{split}
\end{equation}

Factoring out the common terms, we get
\begin{equation}
\small
\begin{aligned}
P(\mathcal{C}_{x}\mid i^* = i)
&= \frac{\exp\!\big(R_{\theta}(x,\tau_i)\big)}{Z_{\theta}(x)}
   \underbrace{\left(\prod_{k=0}^{K}\pi_{0}(\tau_k\mid x)\right)}_{\text{Common Base Measure }\Omega(x)} .
\end{aligned}
\end{equation}

The term $\Omega(x)/Z_{\theta}(x)$ depends only on the set $\mathcal{C}_x$ and the prompt, and is independent of the index $i$. 
Substituting this into the posterior, this common factor appears both in the numerator as well as in every summand of the denominator, thus canceling perfectly yielding
\begin{equation}
\begin{aligned}
P(i^* = i \mid \mathcal{C}_{x}, x)
&= \frac{\frac{\Omega(x)}{Z_{\theta}(x)}\exp\!\big(R_{\theta}(x,\tau_i)\big)}
        {\sum_{m=0}^{K}\frac{\Omega(x)}{Z_{\theta}(x)}\exp\!\big(R_{\theta}(x,\tau_m)\big)} \\
&= \frac{\exp\!\big(R_{\theta}(x,\tau_i)\big)}
        {\sum_{m=0}^{K}\exp\!\big(R_{\theta}(x,\tau_m)\big)} .
\end{aligned}
\end{equation}
which is independent of both the partition function $Z_\theta(x)$ as well as the baseline density $\pi_0$.
\hfill $\square$

\paragraph{Corollary A.2 (Identifiability).} 
\label{App: Cor A.2}
\textit{The reward function $R_{\theta}(x, \tau)$ is identifiable up to a prompt-dependent additive constant. That is, $R$ and $R'$ are observationally equivalent if and only if $R'(x,\tau) = R(x,\tau) + f(x)$.}

\textit{Proof.} First, consider sufficiency. Let $\tilde{R}(x, \tau) = R_{\theta}(x, \tau) + f(x)$ for any arbitrary function $f(x)$. Substituting this into the posterior from Theorem A.1:
\begin{equation}
\begin{aligned}
P\!\left(i^*=i \mid \mathcal{C}_x, x\right)
&= \frac{\exp\!\big(\tilde{R}(x,\tau_i)\big)}
     {\sum_{j=0}^{K}\exp\!\big(\tilde{R}(x,\tau_j)\big)} \\
&= \frac{\exp\!\big(R_{\theta}(x,\tau_i)+f(x)\big)}
     {\sum_{m=0}^{K}\exp\!\big(R_{\theta}(x,\tau_m)+f(x)\big)} \\
&= \frac{\exp\!\big(f(x)\big)\exp\!\big(R_{\theta}(x,\tau_i)\big)}
        {\exp\!\big(f(x)\big)\sum_{m=0}^{K}\exp\!\big(R_{\theta}(x,\tau_m)\big)} \\
&= P\!\left(i^*=i \mid \mathcal{C}_x, x\right).
\end{aligned}
\end{equation}

Thus, the posterior remains invariant. 

Conversely, for necessity, if two reward functions induce identical posteriors for all possible contrastive sets, their log-odds must match:
\begin{equation*}
    \begin{aligned}
        & \log \frac{P\!\left(i^*=i \mid \mathcal{C}_x, x\right)}{P\!\left(i^*=j \mid \mathcal{C}_x, x\right)} \\
        &= R(x, \tau_i) - R(x, \tau_j) = R'(x, \tau_i) - R'(x, \tau_j).
    \end{aligned}
\end{equation*}
This implies that $R'(x, \tau) - R(x, \tau) = \text{const}$ (w.r.t. $\tau$), which defines the function $f(x)$.
Thus, prompt-dependent additive shifts characterize the complete equivalence class. Consequently, while global reward magnitudes are not unique, relative reward differences and gradient-based attributions (since $\nabla_\tau f(x) = 0$) are uniquely identified.

\section{C-IRL Implementation Details}
\label{app:cirl_implementation}

\subsection{C-IRL Algorithm}
\label{app:algorithm}

Algorithm~\ref{alg:cirl} presents the complete C-IRL training procedure. The algorithm consists of two phases: (1) dataset construction, where we collect trajectory samples from both the baseline and aligned policies, and (2) contrastive training, where we optimize the reward network to distinguish aligned trajectories from baseline trajectories.

\begin{algorithm}[H]
\caption{Contrastive Inverse Reinforcement Learning (C-IRL)}
\label{alg:cirl}
\begin{algorithmic}[1]
\Require Baseline policy $\pi_0$, aligned policy $\pi_E$, prompt dataset $\mathcal{D}$
\Require Number of negatives $K$, learning rate $\eta$, training iterations $T$
\Ensure Trained reward network $R_\theta$

\Statex \textbf{Phase 1: Dataset Construction}
\State Initialize trajectory buffers $\mathcal{B}_E \gets \{\}$, $\mathcal{B}_0 \gets \{\}$
\For{each prompt $x \in \mathcal{D}$}
    \State Sample $\tau^+ \sim \pi_E(\cdot | x)$ \Comment{Expert trajectory}
    \State Sample $\{\tau_j^-\}_{j=1}^{K} \sim \pi_0(\cdot | x)$ \Comment{Baseline trajectories}
    \State $\mathcal{B}_E \gets \mathcal{B}_E \cup \{(x, \tau^+)\}$
    \State $\mathcal{B}_0 \gets \mathcal{B}_0 \cup \{(x, \tau_j^-)\}_{j=1}^{K}$
\EndFor

\Statex
\Statex \textbf{Phase 2: Contrastive Training}
\State Initialize reward network $R_\theta$ from pretrained encoder
\State Initialize optimizer (e.g., AdamW with weight decay)
\For{$t = 1, \ldots, T$}
    \State Sample minibatch of prompts $\{x_b\}_{b=1}^{B} \subset \mathcal{D}$
    \For{each prompt $x_b$ in minibatch}
        \State Retrieve expert trajectory: $\tau_b^+ \gets \mathcal{B}_E[x_b]$
        \State Sample $K$ negatives: $\{\tau_{b,j}^-\}_{j=1}^{K} \subset \mathcal{B}_0[x_b]$
        \State Compute $r_b^+ \gets R_\theta(x_b, \tau_b^+)$
        \State Compute $r_{b,j}^- \gets R_\theta(x_b, \tau_{b,j}^-)$ for $j = 1, \ldots, K$
    \EndFor
    
    \Statex \Comment{InfoNCE loss and gradient update}
    \State $\mathcal{L} \gets -\frac{1}{B} \sum_{b=1}^{B} \log \dfrac{\exp(r_b^+)}{\exp(r_b^+) + \sum_{j=1}^{K} \exp(r_{b,j}^-)}$
    \State $\theta \gets \theta - \eta \nabla_\theta \mathcal{L}$
\EndFor
\State \Return $R_\theta$
\end{algorithmic}
\end{algorithm}

\section{SAE Training Algorithm}
\label{app:sae_algorithm}

Algorithm~\ref{alg:sae} presents the SAE training procedure for reward decomposition.

\begin{algorithm}[H]
\caption{Sparse Autoencoder for Reward Decomposition}
\label{alg:sae}
\begin{algorithmic}[1]
\Require Trained reward network $R_\theta$ with linear head $(w_{\mathrm{head}}, b_{\mathrm{head}})$
\Require Dataset of prompt-response pairs $\mathcal{D} = \{(x_i, \tau_i)\}_{i=1}^{N}$
\Require Hidden dim $d$, dictionary size $M$, sparsity $k$, consistency weight $\lambda$
\Require Training iterations $T$, learning rate $\eta$, auxiliary coefficient $\alpha$
\Ensure Trained SAE, feature coefficients $\{\omega_i\}_{i=1}^{M}$, constant $C$

\Statex \textbf{Phase 1: Activation Extraction}
\State Initialize activation buffer $\mathcal{H} \gets \{\}$
\For{each $(x, \tau) \in \mathcal{D}$}
    \State $h \gets R_\theta^{\mathrm{penult}}(x, \tau)$ \Comment{Penultimate layer activation}
    \State $\mathcal{H} \gets \mathcal{H} \cup \{h\}$
\EndFor

\Statex
\Statex \textbf{Phase 2: Initialization}
\State $b_{\mathrm{dec}} \gets \text{GeometricMedian}(\mathcal{H})$
\State Initialize $W_{\mathrm{dec}} \in \mathbb{R}^{d \times M}$ with unit-norm orthogonal columns
\State $W_{\mathrm{enc}} \gets W_{\mathrm{dec}}^\top$, \quad $b_{\mathrm{enc}} \gets \mathbf{0}$
\State $\bar{f}_i \gets 0$ for $i = 1, \ldots, M$ \Comment{Feature frequency tracker}

\Statex
\Statex \textbf{Phase 3: Training}
\For{$t = 1, \ldots, T$}
    \State Sample minibatch $\{h_b\}_{b=1}^{B} \subset \mathcal{H}$
    
    \For{$b = 1, \ldots, B$}
        \State $z_b \gets W_{\mathrm{enc}}(h_b - b_{\mathrm{dec}}) + b_{\mathrm{enc|}|}$ \Comment{Eq.~\ref{eq:sae_encode}}
        \State $f_b \gets \text{TopK}(\text{ReLU}(z_b), k)$ \Comment{Sparse features}
        \State $\hat{h}_b \gets W_{\mathrm{dec}} f_b + b_{\mathrm{dec}}$ \Comment{Reconstruction}
    \EndFor
    
    \Statex \Comment{Losses (Eq.~\ref{eq:sae_loss})}
    \State $\mathcal{L}_{\mathrm{recon}} \gets \frac{1}{B} \sum_{b=1}^{B} \| h_b - \hat{h}_b \|_2^2$
    \State $\mathcal{L}_{\mathrm{reward}} \gets \frac{1}{B} \sum_{b=1}^{B} \left( w_{\mathrm{head}}^\top h_b - w_{\mathrm{head}}^\top \hat{h}_b \right)^2$
    
    \Statex \Comment{Dead feature prevention}
    \State $\bar{f}_i \gets 0.99 \cdot \bar{f}_i + 0.01 \cdot \frac{1}{B} \sum_{b=1}^{B} \mathbb{I}[f_{b,i} > 0]$ for all $i$
    \State $\alpha_t \gets \alpha$ if $t < 0.9T$, else linearly anneal to $0$
    \State $\mathcal{L}_{\mathrm{aux}} \gets \alpha_t \sum_{i=1}^{M} \max(0, \frac{k}{2M} - \bar{f}_i)^2$
    
    \Statex \Comment{Update}
    \State $\mathcal{L}_{\mathrm{tot}} \gets \mathcal{L}_{\mathrm{recon}} + \lambda \, \mathcal{L}_{\mathrm{reward}} + \mathcal{L}_{\mathrm{aux}}$
    \State Update $(W_{\mathrm{enc}}, b_{\mathrm{enc}}, W_{\mathrm{dec}}, b_{\mathrm{dec}})$ via AdamW
    \State $\mathbf{d}_i \gets \mathbf{d}_i / \|\mathbf{d}_i\|_2$ for $i = 1, \ldots, M$ \Comment{Normalize decoder}
\EndFor

\end{algorithmic}
\end{algorithm}

\section{Permutation Test for Feature Significance}
\label{app:selection statistics}

To ensure that the identified hacking features represent genuine systematic patterns rather than sampling artifacts, we employ a permutation-based hypothesis test. Under the null hypothesis $H_0$, the feature contributions on hacked examples are drawn from the same distribution as the general population, implying no systematic elevation of feature $i$ on hacked examples.

The test proceeds as follows. For each feature $i$, we first compute the observed hacking contribution score $H_i^{\mathrm{obs}}$ using Equation~\ref{eq:hacking_score}. We then generate $B$ permutation samples, typically $B = 1000$, by randomly sampling $|\mathcal{D}_{\mathrm{hack}}|$ examples from the SAE training distribution $\mathcal{D}_{\mathrm{SAE}}$ and computing the hacking score $H_i^{(b)}$ on each random sample. The $p$-value is computed as the fraction of permutations where the permuted score meets or exceeds the observed score: $p_i = \frac{1}{B} \sum_{b=1}^{B} \mathbf{1}[ H_i^{(b)} \geq H_i^{\mathrm{obs}} ]$.

Since we test $M$ features simultaneously, we must correct for multiple comparisons to control the false discovery rate. We apply the Benjamini-Hochberg procedure at level $\alpha = 0.05$. The $p$-values are sorted in ascending order as $p_{(1)} \leq p_{(2)} \leq \cdots \leq p_{(M)}$, and we find the largest $k$ such that $p_{(k)} \leq \frac{k}{M} \alpha$. All features with $p_i \leq p_{(k)}$ are deemed statistically significant. The final set of hacking features $\mathcal{S}_{\mathrm{hack}}$ is obtained by intersecting the statistically significant features with the directionality filter $\mathcal{S}_{\mathrm{positive}} = \{i : \bar{H}_i > 0\}$.

\section{Implementation Details for Mitigation}
\label{app:mitigation_implementation}

This section provides complete implementation specifications for all mitigation methods.



\subsection{PPO Configuration}

For Methods A, B, and C, which require reinforcement learning, we use the following PPO configuration. The learning rate is set to $1 \times 10^{-5}$ with linear warmup over the first 10\% of training steps. The KL coefficient $\beta$ is initialized at 0.1 with adaptive adjustment to maintain a target KL divergence. The PPO clip range is set to 0.2, the value function coefficient to 0.5, and the entropy coefficient to 0.01 to encourage exploration. We use a batch size of 64 with mini-batches of size 8 and perform 4 PPO epochs per batch. Gradient accumulation over 4 steps yields an effective batch size of 256. Gradients are clipped to a maximum norm of 1.0 for stability.

\subsection{Distillation Configuration}

For Method D, we generate $N = 16$ candidates per prompt using temperature 0.8 and top-$p$ of 0.95. The penalty strength $\eta$ in the surgical reward is set to 1.0. The minimum surgical reward threshold $\tau_{\mathrm{min}}$ is set to 0.0, meaning we only require the selected response to have non-negative surgical reward. Fine-tuning uses a learning rate of $2 \times 10^{-5}$ for 3 epochs with batch size 32. This configuration is significantly simpler than the PPO-based methods and requires no reward model in the training loop.

\subsection{Computational Requirements}

The computational requirements vary substantially across methods. Methods A, B, and C require approximately 20-50 GPU-hours on 8$\times$ A100 GPUs for a 7B parameter model, with peak memory usage of 40-80 GB depending on batch size and sequence length. The PPO infrastructure adds significant overhead. Method D requires only 5-10 GPU-hours with peak memory of 20-40 GB, as it uses standard supervised fine-tuning without online RL. The candidate generation step can be parallelized efficiently across GPUs. For practitioners with limited computational resources, Method D offers the most favorable trade-off between effectiveness and efficiency.

\section{Ablation Studies: Full Results}
\label{app:ablation_cirl}

This section provides detailed ablation experiments examining the impact of key hyperparameters and design choices on C-IRL's reconstruction fidelity and policy recovery performance.

\subsection{Number of Negative Samples ($K$)}
\label{app:ablation_k}

The InfoNCE objective (Equation~\ref{eq:infonce}) contrasts each positive response from $\pi_E$ against $K$ negative responses sampled from $\pi_0$. Increasing $K$ provides a tighter lower bound on mutual information and reduces gradient variance, but incurs higher computational cost.

\paragraph{Setup.} We train C-IRL with $K \in \{1, 2, 4, 8, 16\}$ while holding all other hyperparameters fixed (50K contrastive pairs, 98M parameter reward network, Llama-2-7B backbone). We measure Spearman correlation with $R_{\text{GT}}$, pairwise ranking accuracy, forward KL divergence of the recovered policy, gradient variance during training, and wall-clock training time.

\paragraph{Results.} Table~\ref{tab:rq1_ablation_k} shows that performance improves from $K{=}1$ to $K{=}4$ and then largely saturates. Ranking accuracy reaches 87.8\% at $K{=}4$, while increasing to $K{=}16$ yields only a marginal gain (+2.4\%) at approximately $4\times$ the computational cost. Gradient variance decreases monotonically with $K$, consistent with the theoretical analysis of InfoNCE~\citep{oord2018representation}. We use $K{=}4$ as the default throughout our experiments.

\begin{table*}[t]
\centering
\caption{\textbf{Effect of the number of negative samples $K$.} Performance saturates around $K=4$, providing the best trade-off between reconstruction quality and computational cost.}
\label{tab:rq1_ablation_k}
\begin{tabular}{@{}cccccc@{}}
\toprule
$K$ & \textbf{Spearman} $\rho$ & \textbf{Rank Acc.} & \textbf{Forward KL} & \textbf{Grad. Var.} & \textbf{Train Time} \\
\midrule
1 & 0.86 {\scriptsize $\pm$ .02} & 85.2\% & 0.032 & 1.0$\times$ & 1.0$\times$ \\
2 & 0.87 {\scriptsize $\pm$ .01} & 86.1\% & 0.028 & 1.4$\times$ & 1.9$\times$ \\
4 & \textbf{0.89} {\scriptsize $\pm$ .01} & \textbf{87.8\%} & \textbf{0.021} & 1.8$\times$ & 3.7$\times$ \\
8 & 0.89 {\scriptsize $\pm$ .01} & 87.9\% & 0.020 & 2.6$\times$ & 7.2$\times$ \\
16 & 0.90 {\scriptsize $\pm$ .01} & 88.2\% & 0.019 & 3.1$\times$ & 14.1$\times$ \\
\bottomrule
\end{tabular}
\end{table*}

\subsection{Contrastive Dataset Size}
\label{app:ablation_data}

The contrastive dataset consists of paired responses $(x, \tau^E, \tau^0)$ where $\tau^E \sim \pi_E(\cdot|x)$ and $\tau^0 \sim \pi_0(\cdot|x)$. Larger datasets provide better coverage of the behavioral shift between the two policies but require more data collection and training time.

\paragraph{Setup.} We train C-IRL with dataset sizes $\in \{5\text{K}, 10\text{K}, 25\text{K}, 50\text{K}, 100\text{K}\}$ contrastive pairs while holding other hyperparameters fixed ($K{=}4$, 98M parameter reward network, Llama-2-7B backbone).

\paragraph{Results.} Table~\ref{tab:rq1_ablation_data} shows that reconstruction quality improves substantially as dataset size grows from 5K to 50K samples (Spearman $\rho$: 0.79$\rightarrow$0.89; forward KL: 0.58$\rightarrow$0.21), after which gains diminish. This suggests that approximately 50K contrastive pairs provide sufficient coverage of the behavioral shift between $\pi_0$ and $\pi_E$ for reliable reward recovery. We use 50K pairs as the default.

\begin{table*}[t]
\centering
\caption{\textbf{Effect of contrastive dataset size.} Performance saturates around 50K contrastive pairs.}
\label{tab:rq1_ablation_data}
\begin{tabular}{@{}cccc@{}}
\toprule
\textbf{Dataset Size} & \textbf{Spearman} $\rho$ & \textbf{Rank Acc.} & \textbf{Forward KL} \\
\midrule
5K & 0.79 {\scriptsize $\pm$ .03} & 80.2\% & 0.058 \\
10K & 0.84 {\scriptsize $\pm$ .02} & 84.1\% & 0.038 \\
25K & 0.87 {\scriptsize $\pm$ .01} & 86.5\% & 0.26 \\
50K & \textbf{0.89} {\scriptsize $\pm$ .01} & \textbf{87.8\%} & \textbf{0.021} \\
100K & 0.89 {\scriptsize $\pm$ .01} & 88.1\% & 0.020 \\
\bottomrule
\end{tabular}
\end{table*}

\subsection{Reward Network Capacity}
\label{app:ablation_capacity}

Table~\ref{tab:nn_size} shows that reconstruction fidelity improves markedly as capacity grows from 24M to approximately 100M parameters, but then saturates: the 98M and 210M models achieve essentially identical Spearman correlation (0.89) and forward KL (0.021 vs 0.020). This suggests that once $\widehat{R}_\theta$ is sufficiently expressive, the limiting factor becomes the information content and coverage of the behavioral contrasts rather than model capacity. We use the 98M configuration as the default.

\begin{table*}[t]
\centering
\caption{\textbf{Effect of reward network capacity.} Performance saturates around 100M parameters.}
\label{tab:nn_size}
\begin{tabular}{@{}ccccc@{}}
\toprule
\textbf{Hidden Dim.} & \textbf{Depth} & \textbf{Params} & \textbf{Spearman} $\rho$ & \textbf{Forward KL} \\
\midrule
512 & 2 & 24M & 0.78 & 0.52 \\
1024 & 4 & 46M & 0.85 & 0.35 \\
2048 & 8 & 98M & \textbf{0.89} & \textbf{0.21} \\
4096 & 12 & 210M & 0.89 & 0.20 \\
\bottomrule
\end{tabular}
\end{table*}

\subsection{Backbone Model Generalization}
\label{app:ablation_backbone}

C-IRL reconstructs $\widehat{R}_\theta$ using features extracted from an LLM backbone, which serves as the semantic representation space for contrastive learning. While the C-IRL objective is model-agnostic, the choice of backbone can affect representation quality and the geometry of the learning problem.

\paragraph{Setup.} We evaluate C-IRL on three widely used backbones: Llama-2-7B~\citep{touvron2023llama}, Mistral-7B~\citep{jiang2023mistral}, and Llama-3-8B~\citep{llama3}. For each backbone, we use the same hyperparameters ($K{=}4$, 50K pairs, 98M reward network) and measure reconstruction fidelity and policy recovery.

\paragraph{Results.} Table~\ref{tab:backbone} shows that C-IRL achieves comparable reward agreement and policy recovery across all backbones, with Spearman $\rho$ ranging from 0.88 to 0.89 and forward KL from 0.20 to 0.23. This demonstrates that C-IRL is not tied to a particular architecture and can be applied with different foundation models.

\begin{table*}[t]
\centering
\caption{\textbf{C-IRL performance across different LLM backbones.} The method generalizes across architectures.}
\label{tab:backbone}
\begin{tabular}{@{}lcccc@{}}
\toprule
\textbf{Backbone} & \textbf{Params} & \textbf{Spearman} $\rho$ & \textbf{Rank Acc.} & \textbf{Forward KL} \\
\midrule
Llama-2-7B & 7B & 0.89 & 87.8\% & 0.021 \\
Mistral-7B & 7B & 0.88 & 87.2\% & 0.023 \\
Llama-3-8B & 8B & 0.89 & 88.1\% & 0.020 \\
\bottomrule
\end{tabular}
\end{table*}

\begin{table}[h]
\centering
\caption{\textbf{Default C-IRL hyperparameters.}}
\label{tab:default_hyperparams}
\begin{tabular}{@{}ll@{}}
\toprule
\textbf{Hyperparameter} & \textbf{Value} \\
\midrule
Number of negatives ($K$) & 4 \\
Contrastive dataset size & 50K pairs \\
Reward network hidden dim & 2048 \\
Reward network depth & 8 layers \\
Reward network params & 98M \\
Learning rate & $1 \times 10^{-5}$ \\
Batch size & 64 \\
Training epochs & 3 \\

\bottomrule
\end{tabular}
\end{table}
\end{document}